# Synthetic Data for Object Classification in Industrial Applications

August Baaz[1], Yonan Yonan[1], Kevin Hernandez-Diaz[1] [a]
Fernando Alonso-Fernandez[1] [b], Felix Nilsson[2]
[1]*School of Information Technology (ITE), Halmstad University, Sweden*
[2] *HMS Industrial Networks AB, Halmstad, Sweden*
augbaa19@student.hh.se, yonan.adnan707@gmail.com, {kevher, feralo}@hh.se, fenil@hms.se

Abstract: One of the biggest challenges in machine learning is data collection. Training data is an important part since it determines how the model will behave. In object classification, capturing a large number of images per object and in different conditions is not always possible and can be very time-consuming and tedious. Accordingly, this work explores the creation of artificial images using a game engine to cope with limited data in the training dataset. We combine real and synthetic data to train the object classification engine, a strategy that has shown to be beneficial to increase confidence in the decisions made by the classifier, which is often critical in industrial setups. To combine real and synthetic data, we first train the classifier on a massive amount of synthetic data, and then we fine-tune it on real images. Another important result is that the amount of real images needed for fine-tuning is not very high, reaching top accuracy with just 12 or 24 images per class. This substantially reduces the requirements of capturing a great amount of real data.

## 1 INTRODUCTION

Popularized since 2015, Industry 4.0 (Xu et al., 2021) refers to integrating Computer Vision (CV), Artificial Intelligence (AI), Machine Learning (ML), the Internet of Things (IoT), and cloud computing into industrial processes. Some significant changes of industry 4.0 are increased automation, self-optimization, and predictive maintenance. For example, object detection and image classification could significantly benefit industrial scenarios. Models need training data to learn, and the quality and quantity of such data is the most crucial part to obtain a reliable model. However, collecting data can be challenging and costly.

This research explores methods to minimize the data collection needed to train object recognition and classification. We aim at developing a system to recognize industrial products using a camera. It could monitor production lines and reduce human repetitive workload for tasks such as sorting, inventory keeping, and quality control. We use ResNet50 (He et al., 2016) Convolutional Neural Network (CNN) as classification architecture in conjunction with methods to reduce the amount of data needed, exploring possibilities other than manually collecting a large number of images per class. Our most important contribution is the use of synthetic data rendered with a game engine. Synthetic data is then combined with real data, demonstrating by experiments that the classification network not only keeps a good accuracy but increases its confidence in classifying the different objects.

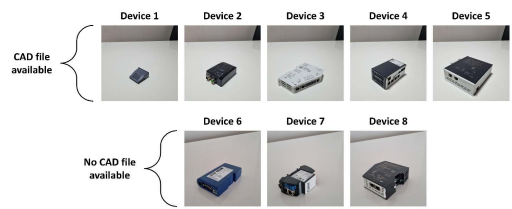

Figure 1: Target objects to be classified. **Device 1**: CompactCom M40 Module EtherNet/IP IIoT Secure. **D2**: Wireless Bridge II Ethernet. **D3**: Communicator PROFINET IO-Device Modbus TCP server. **D4**: Edge Gateway with Switch. **D5**: X-gateway Modbus Plus Slave PROFINET-IRT Device. **D6**: Communicator PROFINET-IRT. **D7**: Edge Essential Sequence. **D8**: Anybus PROFINET to .NET Bridge. All devices can be found at www.anybus.com

This project is a collaboration of Halmstad University with HMS Networks AB in Halmstad. HMS makes products that enable industrial equipment to communicate over various industrial protocols (HMS, 2022). They explore emerging technologies, and one crucial technology is AI, where they want to exam-ine different applications of AI and vision technologies, e.g. (Nilsson et al., 2020), which may be part of future products. As shown in Figure 1, HMS products have simple shapes, although the system is potentially applicable to other products in the industry where sorting and flow control are needed.

---

[a] https://orcid.org/0000-0002-9696-7843
[b] https://orcid.org/0000-0002-1400-346X

## 2 RELATED WORKS

### 2.1 Object Classification

Image classification is a well-known CV field applied to various tasks (Al-Faraj et al., 2021). A CNN-based visual sorting system can be used in an inventory or a warehouse where items lack other tokens, such as a damaged barcode or unreadable tag (Wang et al., 2020). Tailored to retail, (Femling et al., 2018) identified fruit and vegetables with a video-camera attachable to a scale, which could aid or relieve customers and cashiers of navigating through a menu.

A visual-based system is also beneficial for quality control in manufacturing. An operator can get tired after many quality checks and thus misclassify products. To avoid that, (Hachem et al., 2021) implemented ResNet50 for automatic quality control.

In recycling, waste has to be sorted to be recycled properly. This has been studied in (Gyawali et al., 2020) using CNNs, achieving an accuracy of 87%. Similarly, (Persson et al., 2021) developed a method to short plastics from Waste from Electrical and Electronic Equipment (WEEE).

Surveillance is another field. (Jung et al., 2017) detected (using YOLOv4) and classified (using ResNet) various vehicle types, including cars, bicycles, buses and motorcycles. Similarly, (Svanström et al., 2021) developed a drone detector via sensor fusion, being able to distinguish drones from other typical objects, such as airplanes, helicopters, or birds.

### 2.2 Synthetic Data

Ship classification from overhead imagery is a largely unsolved problem in the maritime domain. The main issue is the lack of ground truth data. (Ward et al., 2018) addressed this by building a large-scale synthetic dataset using the Unity game engine and 3D models of ships, demonstrating that synthetic data increases performance dramatically while reducing the amount of real data required to train the models.

For car surveillance, game engines such as Grand Theft Auto V are an excellent way to generate real-looking synthetic images (Richter et al., 2016).

(Tremblay et al., 2018) applied this, achieving an average precision of 79%, which is similar to (Jung et al., 2017) with real data. Thus, it is safe to say that similar results can be achieved training with synthetic data, with the advantage that it is far easier to collect. Also, in (Tremblay et al., 2018), the results of synthetic data far exceed the results of real data after fine-tuning with as few as 100 images.

This work is about sorting industrial products. We can assume that they have a CAD file used in their manufacturing process. 3D scanning is also an effective way. If the object cannot be 3D scanned and does not have a CAD file, Generative Adversarial Networks (GANs) can be used. GANs artificially create similar data using a discriminator that checks if the feature distribution of the generated data looks close to the real data. Some notable GANs are StyleGAN for face generation (Karras et al., 2021) and CycleGAN (Zhu et al., 2017), which allows translating an image from one domain to another (e.g. indoor to outdoor, summer to winter, etc.)

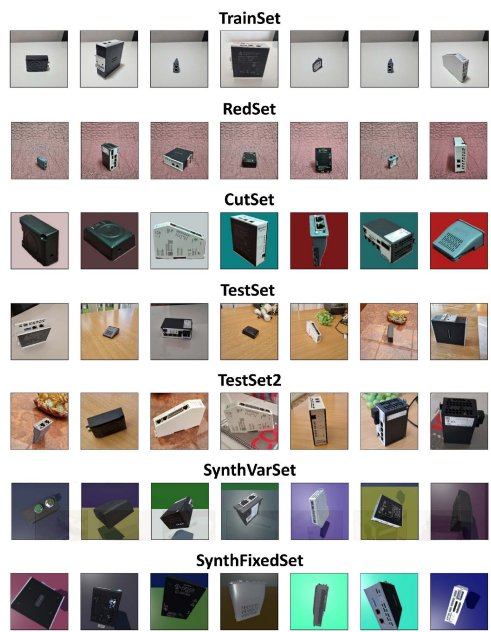

Figure 2: Example of images from the different datasets.

## 3 METHODOLOGY

### 3.1 Data Acquisition and Synthesis

An overview of the different datasets created for this work is given in Table 1. Several HMS products are chosen as target objects (Figure 1). They are

Table 1: Datasets created for this work. The indicated devices are shown in Figure 1.

**Initial stage of our research**

| Name | Data | Devices | Classes | Images/class | Scale | Rotation | Notes |
|---|---|---|---|---|---|---|---|
| TrainSet | Real | 3 4 5 6 7 8 | 6 | 96 | Random | Random | Light on/off |
| TestSet | Real | 3 4 5 6 7 8 | 6 | 30 | Random | Random | Cluttered background |

**Later stage of our research**

| Name | Data | Devices | Classes | Images/class | Scale | Rotation | Notes |
|---|---|---|---|---|---|---|---|
| TrainSet | Real | 1 2 3 4 5 | 5 | 96 | Random | Random | Light on/off |
| RedSet | Real | 1 2 3 4 5 | 5 | 48 | Random | Random | Light on, red background |
| CutSet | Real | 1 2 3 4 5 | 5 | 48 | Fixed | Random | RedSet with fixed scale + transparent background |
| TestSet | Real | 1 2 3 4 5 | 5 | 30 | Random | Random | Cluttered background |
| TestSet2 | Real | 1 2 3 4 5 | 5 | 30 | Fixed | Random | TestSet with fixed scale |
| SynthVarSet | Synth. | 1 2 3 4 5 | 5 | 2000 | Random | Random | Tries to recreate TrainSet conditions |
| SynthFixedSet | Synth. | 1 2 3 4 5 | 5 | 2000 | Fixed | Random | Like SynthVarSet but with fixed scale |

mostly routers and switches for industrial machines that HMS sells. Our research was conducted in two stages. In the initial one, we started to build a dataset of training and test data with devices 3 to 8. However, we did not have accessibility to CAD files of all these products initially used. For this reason, we later employed devices 1 to 5, for which CAD files were available to enable the possibility of creating 3D synthetic data. However, a few experiments were conducted on the initial dataset before switching to the later one, but they were not re-run, so we keep the description of both datasets here. In the experimental section, we will make clear which one is being used in each particular experiment.

Real images from each product were captured using a smartphone with 4K resolution, creating the following datasets (see Figure 2):

- **TrainSet**: using the smartphone on a tripod to simulate a stationary camera, with 96 images per class. Each object is rotated on all sides, with lights on/off to vary the illumination in the room.

- **RedSet**: created exactly like TrainSet, except that it has a red backdrop that can be segmented digitally. This dataset has 48 pictures per class since it was only captured with lights on.

- **CutSet**: created from RedSet by segmenting the background and normalizing the object scale. The segmentation mask is created by applying thresholds to the HSV channels. Morphological opening and closing are also applied to clean artifacts. The background is transparent, allowing addition of different backgrounds as desired to simulate different environments. In our experiments, it is replaced with a random RGB color during runtime.

- **TestSet**: for evaluation, with 30 images per class. It has cluttered backgrounds, and not stationary camera, so the target objects may not be the only objects in the image.

- **TestSet2**: created by cutting the objects of interest of TestSet manually to fill the entire frame.

Using Unity's universal render pipeline (URP), synthetic data is also generated. It takes 16ms to generate one image, or 3750 images/minute, which is a fast and reliable way to generate datasets in minutes. A Unity scene has been built with a room that can be randomized. The camera that renders the scene can be programmed to focus anywhere in the room, and its distance to the object can be varied. The intensity and rotation of the lighting can also be randomized, and the background can be replaced too. Unity offers a library called Perception (Borkman et al., 2021). Using this package, the images for the synthetic datasets are artificially generated. The same seed is provided for every product to generate the same random scene for every object, as can be seen in Figure 3.

A 3D model of the objects has to be obtained for Unity to render images. If the products have a CAD file, they can be converted into 3D models. HMS provided us with all the CAD files for the products. With this, two synthetic datasets are created, having 2000 images per class (see Figure 2). Every synthetic dataset also uses the same seed to generate the same randomness for every object:

- **SynthVarSet**: created with varying distances between the object and the camera to simulate different scaling, and with randomized orientation, light direction and background. This dataset tries to simulate and recreate TrainSet.

- **SynthFixedSet**: is exactly like SynthVarSet with randomized rotations and backgrounds. The difference is that the distance between the objects and the camera is fixed, so that every object fills the frame, thus normalizing the scale.

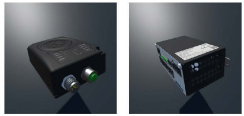

Figure 3: Synthetic images of device2 and device4 generated with the same seed.

## 3.2 System Overview

This research has developed an AI to classify industrial products (Figure 1) with a camera. This is about finding the type of object (class or category) that is appearing in the image. This significantly differs in complexity depending on the specific scenario. For this work, the following limitations are considered: *i)* the camera is stationary, located to one side, and angled towards the table where objects are located; *ii)* the camera is in colour, *iii)* the objects are in focus, *iv)* the table is well-lit, and the objects are visible, and *v)* only one product needs to be identified at a time. The objective is to identify products reliably (measured by accuracy on a test set not seen during training) regardless of their orientation or scale.

The model architecture is based upon ResNet50 pre-trained on ImageNet as a feature extractor. The network is connected to a single fully connected layer with dropout, followed by a five/six neurons layer (the number of classes in our datasets) with softmax activation. The original ResNet50 has two fully connected layers of 4096 each, which makes up for a large portion of the weights (Reddy and Juliet, 2019). Using only a single layer at the end of ResNet50 is entirely arbitrary but common for transfer learning with ResNet. During training, we will test the optimal size of this fully connected layer, as well as the number of early layers that are frozen. Generally speaking, early layers find simple patterns that are general for vision tasks, such as lines or shapes, and they can be kept frozen. On the other hand, later layers find more complex patterns that are specific of each task, (Yosinski et al., 2015). so it is expected to benefit accuracy by re-training these last layers on the task-specific data.

## 4 EXPERIMENTS AND RESULTS

Many parameters affect the training and performance of a machine-learning model like ours. A series of experiments were done to find the best model parameters. The learning rate highly depends on other factors, so we tune it up and down in all tests to do a grid search. The results reported on each sub-section correspond to the learning rate that produces the best numbers. The batch size is kept constant at 64, except in Section 4.1.4, where the experiments demand to change this value. The main evaluation metric employed is Accuracy, given by the fraction between correctly classified trials and the total amount of trials. For a given object (class) to be classified, we also employ *i)* Precision (P), the fraction between True Positives (number of correctly detected objects of the class), and the total amount of trials labeled as belonging to the class, and *ii)* Recall (R), the fraction between True Positives and the total amount of trials that belong to the class. Precision measures the proportion of trials labeled as a given class that are really objects of that class, whereas Recall measures the proportion of objects of a given class that are correctly associated with that class. A single measure summarizing P and R is the F1-score, which is their harmonic mean, computed as F1=2×(P×R)/(P+R). Another way is the confusion matrix, a table that provides the model predictions (*x*-axis) against the true prediction of an object (*y*-axis).

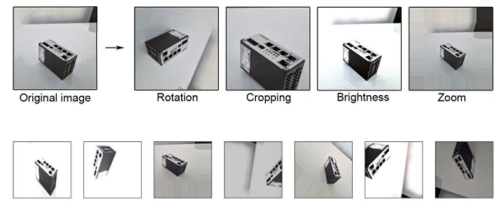

Figure 4: Example of data augmentation methods (top: isolated effect, bottom: combined effect on a single image).

### 4.1 Finding the Best Configuration

#### 4.1.1 Data Augmentation

We first test different data augmentation methods (Figure 4), to test if they allow to combat over-fitting and help the models to better generalize against light and camera changes. These data augmentation experiments are the only ones carried out on the initial dataset with six classes that we gathered (Table 1, top). In all the remaining sections, the later dataset with five classes is employed. Because of that, the test results cannot be compared directly, but we believe that the conclusions of this sub-section, i.e. using data augmentation, are still valid.

Experiments of this sub-section are done on real data, using TrainSet/TestSet as train/test sets, with 80% of TrainSet used for actual training and 20% for validation to stop training. Rotation (360°), cropping (0-30% in all sides), brightness change (50-120%), and zooming (100-150%) are used. For these experiments, the network is left with 32 unfrozen layers at the end and a fully connected layer (before softmax) of 256 elements. The obtained accuracy with-out/with data augmentation is 73.9/84.4%. Without data augmentation, most misclassified images are objects that appear far away (small scale). This suggests that zooming may have a significant effect on model performance. It is well known that CNNs often struggle to identify objects on different scales (Liu et al., 2018). However, the model was trained with all data augmentation methods applied together, so the effect of individual changes was not explored. Given these results, all subsequent models in the project were trained with data augmentation

Table 2: Left: effect of changing the end layer size (unfrozen layers set to 32). Right: effect of changing the number of unfrozen layers (end layer size set to 128).

(unfrozen layers=32)

| End layer size | Accuracy |
|---|---|
| 64 | 63,3 |
| 128 | 75,3 |
| 256 | 70,7 |
| 512 | 61,3 |

(end layer size=128)

| Unfrozen layers | Accuracy |
|---|---|
| 32 | 75,3 |
| 48 | 62,7 |
| 64 | 72,7 |
| 96 | 76,7 |
| 128 | 53,3 |

### 4.1.2 ResNet Model Setup

Here, we test the optimal number of unfrozen layers left at the beginning of ResNet50, and the size of the fully connected layer. Experiments are done on real data, using TrainSet/TestSet as train/test sets and 80/20% for actual training/validation. The more layers are left unfrozen, the more a network is prone to over-fitting if few training data is available, while, while too few unfrozen layers may make the model converge towards a lower accuracy. The size of the end layer also has an impact on the network training time and accuracy. Starting with 32 unfrozen layers and an end layer size of 512, we first reduce the size of the end layer by a factor of 2 (Table 2, left). A too high-end layer size is seen to negatively affect accuracy. The model shows a significant improvement by decreasing the layer size up to 128 and beyond that, accuracy decreases again. We then set the size of the end layer to 128 and increase the number of unfrozen layers from 32 by a factor of 2 (Table 2, right). The best result is obtained with 96 layers unfrozen (55% of the network), and going beyond that hurts accuracy. The difference between 96 and 64 layers unfrozen is ∼4%, but the model with 64 unfrozen layers (37% of the network) has been observed to be more consistent and less prone to over-fitting. Thus, the settings of 64 unfrozen layers and an end layer of 128 are identified as the optimal settings of this subsection, which will be used on all subsequent models.

Table 3: Effect of changing dropout.

| Dropout | Accuracy |
|---|---|
| 0% | 83,3 |
| 20% | 60,7 |
| 40% | 67,3 |
| 60% | 65,3 |
| 80% | 76,0 |
| 100% | 43,3 |

### 4.1.3 Dropout

Dropout regularisation can help to make models more general and reduce over-fitting, but too high dropout might make the model converge towards a lower accuracy. We start with a dropout rate of 0% applied after the fully connected layer, and then increased in steps of 20% (Table 3). Experiments are done on real data, using TrainSet/TestSet as train/test sets and 80/20% for actual training/validation. The best result is obtained when no dropout at all is applied, yielding an accuracy of 83.3% on TestSet. One possible explanation of this result is that we are using a feature vector of 128 elements to classify only 5 classes, so dropout is not providing any tangible benefit. This is quite small compared to feature vectors of 2048 or 4096 elements, which are common in CNN architectures, followed by a classification layer of 1000 elements (e.g. in ImageNet).

Table 4: Effect of reducing the size of the training set.

| Images per class | Training images per class | Batch size | Accuracy % |
|---|---|---|---|
| 12 | 10 | 8 | 62,0 |
| 24 | 20 | 16 | 72,7 |
| 48 | 39 | 32 | 73,3 |
| 96 | 77 | 64 | 72,7 |

### 4.1.4 Reduced Training Data

The TrainSet employed in previous subsections has 96 images per class. In an operational industrial system, taking such amount of images per object that needs to be sorted may be inconvenient. In this subsection, we will reduce the size of the training set to 48, 24 and 12 images per class to assess the practicality of capturing fewer images vs its the effect on accuracy. Accuracy is computed on TestSet. As we decrease the amount of images per class, the minibatch size is also reduced, since models tend to learn badly and overfit when the mini-batch size is too large compared with the dataset size. Results are given in Table 4. The change in accuracy between 96, 48 and 24 is negligible, which is positive for our purposes.

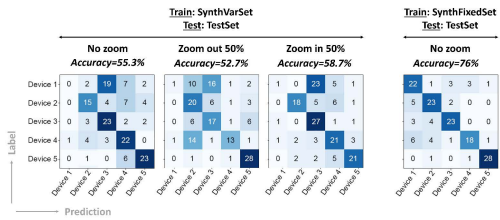

Figure 5: Synthetic model trained on SynthVarSet (left) or SynthFixedSet (right) and tested on TestSet with different scales.

### 4.1.5 Synthetic Data

Products that need to be sorted for the manufacturing industry likely have CAD models. Synthetic data generated from such models can be an alternative to taking pictures of each object. A synthetic dataset can be made many times larger, which can help against over-fitting and make the models to generalize better.

We first train the model on SynthVarSet and compute accuracy results on TestSet (Figure 5, left part). The first sub-element ('no zoom') shows the results with the original SynthVarSet and TestSet datasets. As it can be seen, the model struggles with specific objects (Device 2 and specially Device 1), which turn out to be the smallest objects (see Figure 1). Recall that in these two datasets, the objects appear with variable scale (Figure 2 and Table 1). To check this scale issue, the models were tested again on TestSet with images zoomed out and in by 50%. As it can be seen, this makes that different devices get better or worse under zooming out or in. For example, Device 3 gets better when zooming in, and worse with zooming out, while Device 5 is the opposite. Also, Device 1 (the smallest one) is mis-classified most of the times, no matter in which option. The overall accuracy also increases slightly with zooming in. Since the objects to be detected fill a bigger portion of the image, this may produce that the network is able to detect them better.

The model is then trained on SynthFixedSet (Figure 5, right). SynthFixedSet is generated in a way that the scale is fixed, with objects filling the entire frame. As it can be seen, this training provides the best balanced accuracy among all objects, and the best overall accuracy. Specially Device 1 (the smallest object) is brough to a similar accuracy than the other devices, very likely because now the object occupies a bigger portion of the training images, so the network better can ´see it' when it appears smaller on images of TestSet. The overall accuracy on TestSet is 76%, with the model trained on a synthetic dataset with 2000 images per object. From the experiments of Section 4.1.4, this size could likely be smaller without hurting performance, although such option has not been tested.

Table 5: Effect of fine-tuning the model trained with synthetic images of SynthFixedSet with real images of TrainSet.

| Dataset size | Accuracy % | Improvement |
|---|---|---|
| 96 | 79,3 | 3,3 |
| 48 | 80,7 | 4,7 |
| 24 | 80,0 | 4,0 |
| 12 | 79,3 | 3,3 |

### 4.1.6 Real and Synthetic Data Mix

A model trained on a large synthetic dataset may learn patterns from the synthetic data that do not apply to reality. Fine-tuning the model on a small number of real images may increase the real-world performance. Another approach would be to combine synthetic and real data during a single training round. However, the size difference between our real and synthetic datasets is large, which could prevent the real data from affecting the results much if the images are mixed together.

To test our assumption, the best synthetic model of the previous section (76% accuracy on TestSet) has been retrained again on TrainSet with a lower learning rate. We also carry out the same data reduction of Section 4.1.4 and evaluate a size of 96, 48, 24 and 12 images per class. Results are shown in Table 5. As it can be seen, this fine-tuning provides an extra accuracy improvement, even with a small number of images. The models with 24 and 12 images do not perform much worse than those with 96 and 48, so the synthetic model can be noticeably improved with a small handful of real images.

Table 6: Accuracy of the best three models on different sets.

| Model | TestSet | TestSet2 | RedSet |
|---|---|---|---|
| RealModel | 83,3% | 87,3% | 89,6% |
| SynthModel | 76,0% | 84,7% | 82,1% |
| SynthTuned | 80,7% | 87,3% | 87,1% |

Table 7: Accuracy of the best three models on the TestSet2 dataset when predictions below 70% confidence are discarded. 'Confident proportion' is the amount of images with at least 70% confidence. 'Accuracy of confident' is the accuracy after discarding images with less than 70% confidence.

|  | Confident images | Confident proportion | Accuracy of confident |
|---|---|---|---|
| RealModel | 89 / 150 | 59,3% | 97,8% |
| SynthModel | 88 / 150 | 58,7% | 95,5% |
| SynthTuned | 107 / 150 | 71,3% | 96,3% |

## 4.2 Best Models' Analysis

The best three models from previous sections are brought here for further analysis. We name them as:

- **RealModel**: from dropout experiments of Section 4.1.3, trained on real data (TrainSet).
- **SynthModel**: from Section 4.1.5, trained on synthetic data (SynthFixedSet).
- **SynthTuned**: from Section 4.1.6, trained on synthetic data (SynthFixedSet) and fine-tuned on real data (TrainSet).

Their accuracy on several sets is summarized in Table 6. As it can be observed, performance on TestSet2 is significantly better than on TestSet, with a noticeable improvement with the models that use synthetic data during training. A better performance on TestSet2 can be expected since the objects occupy the entire image, and the cluttered background is removed (see Figure 2). To test which of the two components (object size of background) are affecting the most, we also report results on RedSet, which has objects with variable scale but with a uniform red background. The performance on RedSet appear to be on par with TestSet2, or even better with RealModel, suggesting that eliminating a cluttered background has a more significant impact than normalizing the scale.

Comparatively, SynthTuned (trained on massive synthetic data and fine-tuned on real data) has a performance on-par with RealModel (trained on just real data), so one may question the utility of the employed synthetic data augmentation. However, accuracy does not tell the full story of how well a model performs. Even if an object is identified correctly, the confidence of the classifier in such decision matters. Setting a threshold on confidence is likely how an object classifier would be used in many practical scenarios. To test the effect of such practice, we set a confidence threshold of 70%, so decisions below this threshold are considered 'unsure'. Disregarding objects below this confidence gives the results shown in Table 7. It can be seen that the amount of trials on which the classifier is confident is substantially higher with SynthTuned, revealing an important benefit given by adding synthetic data to the training set. The overall accuracy of the three models is in a similar range (96-98%), but on SynthTuned, such entails a higher number of images that are actually classified correctly.

## 5 CONCLUSIONS

This paper has studied the utility of adding synthetically-generated data to the training of object detection models. One way to artificially create images is by a game engine, with many of the most famous game engines providing libraries specifically for synthetic data (Borkman et al., 2021; Qiu and Yuille, 2016). We focus on industrial production settings, where CAD models are often accessible for manufactured parts, making possible to generate 2D and 3D synthetic images of them. Synthetic images can be rendered very quickly and effortlessly compared to capturing real data, simulating a wide variability of viewpoints, illumination, scale, etc. In addition, the dataset can be auto-labeled, avoiding errors in manual annotation, and the object's position in the image is known at pixel precision. It also offers many more features that can be very hard to obtain with real data, like 3D labeling, segmentation, and human keypoint labels (Borkman et al., 2021).

Here, we train a ResNet50 model pre-trained on ImageNet to classify five different objects (Figure 1). These are objects commercialized by the collaborating partner of this research, HMS Networks AB in Halmstad. A dataset with images of each object type from different viewpoints has also been acquired, both of real and synthetic images (Figure 2) and with different scales, illumination and background (Table 1). We have conducted different experiments to find the optimal setting of the classifier, including data augmentation, number of frozen layers of the network, size of the end layer, or dropout. We also evaluated the impact of reduced training data and the incorporation of synthetic data in the training set. The latter is done by training the classifier first on a massive amount of synthetic data, and then fine-tuning it on real data. Even if the overall accuracy of models trained with synthetic+real data is on-par with models trained with real data only, it has been observed that the addition of synthetic data helps to increase confidence in classification on a significant number of test images. This is an important advantage in industrial settings, where high confidence in the decision is critical in many situations. Another important contribution is that the amount of real data needed to fine-tune the model is not very high to reach top accuracy (just 12-24 images per class), greatly alleviating the need to obtain a substantial number of real images.

Scale or cluttered background has been identified as two relevant issues. When making the objects fill the entire image frame (thus removing the impact of the background) or the background is set to constant on the test data, a performance improvement is observed (Table 6). Training on images where the object fills the entire frame has also been shown to cope with smaller objects in the test data that are otherwise misclassified frequently (Figure 5). This work has considered stationary objects in a relatively simple and well-lit environment. An obvious improvement likely to appear in industrial settings is to allow motion between the camera and the objects, e.g. due to conveyor belts. To do so, further research in the detection and segmentation of moving objects is necessary before presentation to the classifier. Possible solutions to this, depending on the scene complexity, range from a traditional Mean Frame Subtraction (MFS) method to detect moving objects in simple setups where the background remains static for a long time (Tamersoy, 2009) to more elaborated trained approaches such as RetinaNet (Lin et al., 2020) or YOLOv4 (Bochkovskiy et al., 2020) object detectors. The latter is more tolerant to changes in scale, light, multiple objects, and motion, but often they need more training data. This, however, could be addressed with an approach based on synthetic data like the one followed in this paper.

In a warehouse, new products are coming in all the time. In our case, the classifier must be retrained to recognize each new class. Other alternatives for warehouses with many different products would be expanding a classifier without retraining it (Schulz et al., 2020). Using labels attached to products would be another approach to identify objects. For example, (Nemati et al., 2016) employs spiral codes, similar in concept to barcodes, but detectable with any 360-degree orientation (in contraposition to barcodes that need to be properly oriented). However, this would demand manual attachment of labels to the objects.

## ACKNOWLEDGEMENTS

This work has been carried out by August Baaz and Yonan Yonan in the context of their Bachelor Thesis at Halmstad University (Computer Science and Engineering), with the support of HMS Networks AB in Halmstad. Authors Hernandez-Diaz and Alonso-Fernandez thank the Swedish Research Council (VR) and the Swedish Innovation Agency (VINNOVA) for funding their research.